\pdfoutput=1

\documentclass[11pt]{article}

\usepackage{acl}

\usepackage{times}
\usepackage{latexsym}
\usepackage[T1]{fontenc}
\usepackage[utf8]{inputenc}
\usepackage{microtype}
\usepackage{inconsolata}

\usepackage{url}            
\usepackage{booktabs}       
\usepackage{nicefrac}       
\usepackage{multirow}
\usepackage{makecell}

\usepackage[export]{adjustbox}
\usepackage{xspace}
\usepackage{soul}
\usepackage{xcolor} 
\usepackage{changepage,threeparttable} 

\usepackage{amsmath}
\usepackage{amssymb}
\usepackage{amsfonts}
\usepackage{mathtools}

\usepackage{tabularx} 

\usepackage{mathrsfs} 

\newcommand{\methodname}{{\itshape Deep-Thinking}\xspace}

\definecolor{darkred}{rgb}{0.6,0.0,0.0}
\definecolor{darkgreen}{rgb}{0,0.50,0}
\definecolor{lightblue}{rgb}{0.0,0.42,0.91}
\definecolor{orange}{rgb}{0.99,0.48,0.13}
\definecolor{grass}{rgb}{0.18,0.80,0.18}
\definecolor{pink}{rgb}{0.97,0.15,0.45}

\usepackage{listings}

\lstdefinelanguage{PythonPlus}[]{Python}{
  morekeywords=[1]{,as,assert,nonlocal,with,yield,self,True,False,None,} 
  morekeywords=[2]{,__init__,__add__,__mul__,__div__,__sub__,__call__,__getitem__,__setitem__,__eq__,__ne__,__nonzero__,__rmul__,__radd__,__repr__,__str__,__get__,__truediv__,__pow__,__name__,__future__,__all__,}, 
  morekeywords=[3]{,object,type,isinstance,copy,deepcopy,zip,enumerate,reversed,list,set,len,dict,tuple,range,xrange,append,execfile,real,imag,reduce,str,repr,}, 
} 

\lstdefinestyle{colorEX}{
  basicstyle=\small\ttfamily,
  backgroundcolor=\color{white}, 
  commentstyle=\color{darkgreen}\slshape,
  keywordstyle=\color{blue}\bfseries,
  keywordstyle=[2]\color{blue}\bfseries,
  keywordstyle=[3]\color{grass},
  stringstyle=\color{darkred},
  emphstyle=\color{pink}\underbar,
}
\title{Iterative Forward Tuning Boosts In-Context Learning in Language Models}

\newcommand\numberthis{\addtocounter{equation}{1}\tag{\theequation}}
\newcommand{\bv}[1]{\textit{\textbf{#1}}}

\author{
\bf
Jiaxi Yang$^{1,2,*}$\footnotemark[3], 
Binyuan Hui$^{3,*}$, 
Min Yang$^{1}$\footnotemark[2], 
Bailin Wang$^{4}$\\
\bf
Bowen Li$^{5}$,
Binhua Li$^{3}$, 
Fei Huang$^{3}$, 
Yongbin Li$^{3}$\footnotemark[2] \\
$^1$ Shenzhen Institute of Advanced Technology, Chinese Academy of Sciences \\
$^2$ University of Chinese Academy of Sciences\\
$^3$ Alibaba Group,
$^4$ MIT CSAIL,
$^5$ Shanghai AI Laboratory \\
\texttt{\{jx.yang, min.yang\}@siat.ac.cn} \\
\texttt{binyuan.hby@alibaba-inc.com}\\
\url{https://github.com/Yangjiaxi/DeepThinking}
}

\begin{document}

\maketitle

\renewcommand{\thefootnote}{\fnsymbol{footnote}}
\footnotetext{$^{*}$ Equal contribution.}
\footnotetext[3]{Work done during an intern at Alibaba Group.}
\footnotetext[2]{Corresponding authors.}

\begin{abstract}
Despite the advancements in in-context learning (ICL) for large language models (LLMs), current research centers on specific prompt engineering, such as demonstration selection, with the expectation that a single iteration of demonstrations processing can generalize effectively to a given test sample. However, this perspective overlooks the potential benefits derived from multiple iterations involving demonstrations, a practice aligning more closely with the iterative decision-making process exhibited by humans, who often learn through analogy.
In this study, we introduce a novel two-stage framework to boost ICL in LLMs. Specifically, our framework delineates the ICL process into two distinct stages: \methodname and test stages. The \methodname stage incorporates a unique attention mechanism, i.e., iterative enhanced attention, which enables multiple rounds of information accumulation. This mechanism operates by manipulating the Key-Value matrices without training, fostering enhanced understanding capabilities in LLMs by ``\textit{thinking}'' demonstrations multiple times. 
We evaluated \methodname across a range of benchmarks and LLMs, showing its superior performance over vanilla ICL methods and its effectiveness in challenging tasks where demonstration selection is infeasible.

\end{abstract}

\section{Introduction}

\begin{figure*}[htbp]
    \centering
    \includegraphics[width=0.7\linewidth]{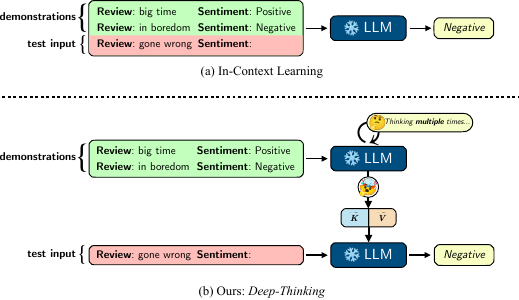}
    \caption{The illustrations of vanilla ICL and our proposed two-stage framework through \methodname. The vanilla ICL method processes demonstrations only once, while our  ``\methodname'' method enables multiple rounds of information accumulation during the reasoning process. 
    }
    \label{fig:intro_highlevel} 
\end{figure*}

Large language models (LLMs), e.g. OpenAI GPTs~\cite{openai2023gpt4}, LLaMA~\cite{touvron2023llama} and Qwen~\cite{bai2023qwen}, demonstrate the mysterious in-context learning (ICL) ability, where LLMs make predictions directly by prepending demonstrations to the original input without updating model parameters. LLMs are expected to learn the patterns hidden in demonstrations and
make predictions accordingly. As illustrated in Figure \ref{fig:intro_highlevel} (a), an LLM can correctly perform inference on an unseen task by conditioning on several demonstrations. The ICL paradigm empowers LLMs to achieve impressive results in various downstream tasks with a few demonstrations, making Language-Model-as-a-Service (LMaaS)~\cite{sun2022blackbox} possible.

Since the performance of ICL is sensitive to specific prompt settings, considerable efforts have been developed to improve the performance of ICL by refining the prompt design from different perspectives, such as demonstration selection \cite{liu-etal-2022-makes,li2023finding},
instruction design \cite{wei2021finetuned, ye2023large}, and intermediate chain-of-thought (CoT) reasoning \cite{wei2022chainofthought,zhang2023automatic, lu2023chameleon}.
These methods can facilitate LLMs to reduce inference variance and avoid poor worst-case accuracy to some extent by performing prompt engineering. The working mechanism of ICL also draws a lot of attention.
\citet{dai2022gpt} shed light on the connections between ICL and explicit fine-tuning. Specifically, ICL computes meta-gradients via forward computation, while explicit fine-tuning obtains gradients by back-propagation. A dual form exists between attention and gradient descent-based optimization \cite{dual-form-icml}, directly connecting the test input to demonstrations. 
\citet{Wang2023LabelWA} argue that label words in demonstrations act as anchors, enabling mapping from demonstrations to test input through information aggregation and label propagation. 

However, these studies assume that the models process demonstrations only once (i.e., perform a single forward computation),
which is incoordinate with the human decision-making process by learning from analogy. Humans usually learn from analogy via an iterative thinking process, such as analyzing demonstrations, reflecting on them, and forming abstract concepts. 
The models learned from demonstrations in inference time by ``\textit{thinking for longer}'' or ``\textit{thinking multiple times}'' \cite{schwarzschild2021can}. 
These findings inspire us to ask a question: \textit{\textbf{Can we boost the performance of ICL by learning from demonstrations through several (iterative) forward inferences?}}

In this paper, we propose a two-stage framework to boost the ICL ability in LLMs. Instead of simply concatenating demonstrations and test input together for inference, we decouple the ICL process into a \methodname stage for demonstration training and a test stage, as illustrated in Figure \ref{fig:intro_highlevel} (b).  
In the \methodname stage, we introduce a new attention module that manipulates the updates of Key-Value matrices~\cite{vaswani2017attention} within the Transformer's self-attention~\cite{vaswani2017attention} mechanism. This modification leverages Key-Value matrices as a bridge to change the information flow to accumulate and learn information over multiple forward iterations without any training.
During the test stage, since the concepts contained in demonstrations are already stored in final Key-Value matrices, we only need to feed the test input into the model and utilize the Key-Value cache for inference.
This \methodname strategy is motivated by humans' repeat logical thinking and reasoning process. LLMs are expected to extend their abilities to solve unseen, complex tasks by
``\textit{thinking}'' demonstrations multiple times.

To verify the effectiveness of the proposed \methodname, we initially conduct evaluations on conventional ICL benchmarks across language models of various sizes. The experiments show that \methodname significantly outperforms vanilla ICL in a variety of model sizes and tasks, surpassing previous state-of-the-art (SOTA) methods focused on selecting demonstrations. In addition, we introduce two more challenging benchmarks (i.e., MMLU~\cite{hendryckstest2021} and BBH~\cite{srivastava2022beyond}) and conduct experiments on advanced LLMs, including LLaMA2~\cite{touvron2023llama} and Pythia~\cite{biderman2023pythia}. We argue that on these challenging benchmarks, demonstration selection becomes impractical due to the lack of a potential candidate pool. \methodname obtains a significant advantage over vanilla ICL. 

\begin{figure*}[ht]
    \centering
    \includegraphics[width=0.95\textwidth]{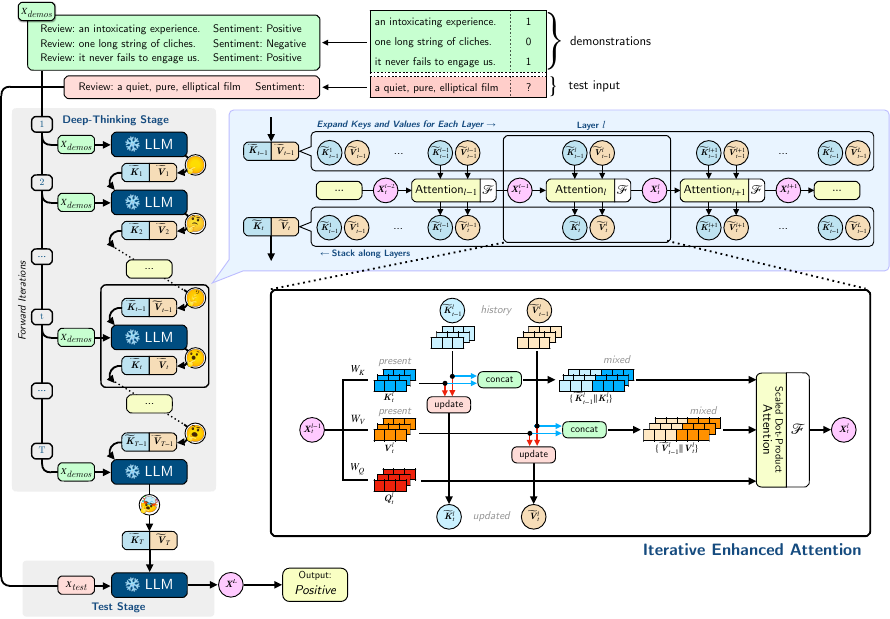}
    \caption{The overview of proposed two-stage ICL framework. It divides the ICL process into \methodname stage and test stage, which take demonstrations and test query as input, respectively. It replaces the vanilla self-attention mechanism with the proposed Iterative Enhanced Attention (IEA). IEA utilizes the Key-Value matrices as bridge of memories, capable of receiving historical (from the previous iteration) memories. It can mix memories with present information to perform attention, and update memories for the next iteration. During testing, predictions are performed using memories that have been refined through multiple iterations. Notably, throughout this process, the LLM parameters remain frozen and no additional parameters are introduced.
    }
    \label{fig:main_methodology} 
\end{figure*}

\section{Preliminaries: In-Context Learning}\label{sec:icl}

This paper focuses on in-context learning tasks.
Formally, given a nature language test input $x_{\textit{test}}$ with a few ($N$-shot) input-output demonstrations $\mathcal{C}_{\textit{demos}} = \{(x_i, y_i)\}_{i=1}^N$, the goal of in-context learning is to predict the label $\hat{y}$ of $x_{\textit{test}}$ from a pre-defined candidate label set $\mathcal{Y} = \{y_1, y_2, ..., y_m\}$ conditioned on $N$ demonstrations.
Given an LLM $\mathcal{M}$ (e.g., a GPT model), the prediction process can be formulated as follows:
\begin{equation}
\hat{y} = \arg\max_{y_j \in \mathcal{Y}} P_{\mathcal{M}}(y_j | \mathcal{C}_{\textit{demos}}, x_{\textit{test}}),
\end{equation}
where $P$ is the output probability of the LLM $\mathcal{M}$. 
Generally, an LLM adopts the Transformer as the backbone, which consists of a stack of several Transformer blocks~\cite{vaswani2017attention}.

\section{Methodology}

In this paper, we propose a two-stage ICL framework that improves performance through multiple forward iterations.
As shown in Figure \ref{fig:main_methodology}, we assign the demonstrations and test input to the \methodname and test stages, respectively, where Key-Value matrices serve as a bridge between the two stages. Next, we describe these two stages in detail.

\subsection{The \methodname Stage}\label{subsec:forward-optim}

\newcommand{\binput}[3]{\bv{#1}_{#2}^{#3}}
\newcommand{\basic}[3]{\bv{#1}_{#2}^{#3}}
\newcommand{\merged}[3]{\widehat{\bv{#1}}{}_{#2}^{#3}}
\newcommand{\update}[3]{\textcolor{blue}{\widetilde{\bv{#1}}{}_{#2}^{#3}}}
\newcommand{\updateB}[3]{\widetilde{\bv{#1}}{}_{#2}^{#3}}

In the \methodname stage, given the demonstrations, we perform multiple forward passes in an iterative way by manipulating the Key-Value matrices in the self-attention~\cite{vaswani2017attention} module.
We use $\bv{X}_t^l$ to denote the output representation of the entire demonstration sequence at layer $l$ and the $t$-th forward pass.
Notably, $\bv{X}_t^l$ receives not only the output $\binput{X}{t}{l-1}$ from the previous Transformer block, but also the Key-Value matrices $\update{K}{t-1}{l}, \update{V}{t-1}{l}$\footnote{Key-Value matrices, represented in \textcolor{blue}{blue}, act as memory carriers throughout the \methodname iterations and serve as inputs in the final test stage.} produced by the same self-attention module at the $(t-1)$-th forward pass.
Accordingly, the Key-Value matrices will be updated as $\update{K}{t}{l}, \update{V}{t}{l}$.

\paragraph{Iterative Enhanced Attention} To handle multiple forward iteration information in \methodname, we proposed a modified attention mechanism, named Iterative Enhanced Attention (IEA). Each block of IEA is illustrated in Figure~\ref{fig:main_methodology}. The information flowing through a block can be observed from both horizontal and vertical processes. The horizontal process represents the calculation of the input parameters in a conventional manner, while the vertical process stands for the manipulation of the Key-Value matrices.
Specifically, the input $\binput{X}{t}{l-1}$ is firstly projected by key, value and query weight matrices, respectively:
\begin{equation}
\scalebox{0.85}{$
\basic{K}{t}{l}=W_K \binput{X}{t}{l-1},\quad\basic{V}{t}{l}=W_V \binput{X}{t}{l-1},\quad\basic{Q}{t}{l}=W_Q \binput{X}{t}{l-1} $}
\label{eq:forward-1}
\end{equation}
where $\basic{K}{t}{l}, \basic{V}{t}{l}$ 
represent the \textbf{present} Key-Value matrices of vanilla self-attention, projected from input $\bv{X}$.
For the horizontal process, we concatenate the \textbf{present} Key-Value matrices with the \textbf{history} Key-Value matrices $\update{K}{t-1}{l}, \update{V}{t-1}{l}$ as the \textbf{mixed} Key-Value to compute attention map and obtain the output $\binput{X}{t}{l}$ of current layer as follows:
\begin{equation}
\scalebox{0.85}{$
\binput{X}{t}{l} = \mathscr{F}(\textbf{Attention}_l(\{ \update{K}{t-1}{l} \Vert \basic{K}{t}{l} \}, \{ \update{V}{t-1}{l} \Vert \basic{V}{t}{l} \}, \basic{Q}{t}{l})) \label{eq:forward-3}$}
\end{equation}
where $\mathscr{F}$ refers to the operations after self-attention, namely the Feed-Forward Network~(FFN)~\cite{vaswani2017attention}, layer normalization~\cite{ba2016layer} and residual connection~\cite{He2015DeepRL}.

Furthermore, the update process is jointly contributed by the \textbf{present} and \textbf{history} Key-Value matrices. From a high-level abstract perspective, the update process can be formalized as follows:
\begin{equation}
\scalebox{0.9}{$
\begin{aligned}
\update{K}{t}{l} &= \textbf{update}(\update{K}{t-1}{l}, \basic{K}{t}{l}) =\eta\basic{K}{t}{l} + (1 - \eta) \update{K}{t-1}{l}\\
\update{V}{t}{l} &= \textbf{update}(\update{V}{t-1}{l}, \basic{V}{t}{l}) =\eta\basic{V}{t}{l} + (1 - \eta) \update{V}{t-1}{l}
\end{aligned}
\numberthis $}
\end{equation}
where $\update{K}{t}{l}$ and $\update{V}{t}{l}$ are \textbf{updated} Key-Value matrices. 
For the update function, we adopt a simple gating mechanism that utilizes the hyper-parameter $\eta$ to control the fusion rate of history and present information.

Modeling demonstrations takes up to $T$ iterations, where the value of $T$ can be predefined by users. After the iterative forward process, we can obtain final \textbf{updated} Key-Value matrices $\update{K}{T}{l}, \update{V}{T}{l}$. By stacking \textbf{updated} Key-Value matrices of all layers in a given LLM, we have
\begin{equation}
\update{K}{T}{}= \{ \update{K}{T}{l} \}_{l=1}^L,\quad\update{V}{T}{} = \{ \update{V}{T}{l} \}_{l=1}^L
\end{equation}
which can be stored statically. $L$ denotes the number of Transformer blocks in an LLM.

\subsection{The Test Stage}

Considering that we now have the Key-Value matrices $\update{K}{T}{}, \update{V}{T}{}$ that have been updated for $T$ iterations, the information contained in them can be regarded as a highly condensed modeling of the demonstrations. The inference process can be performed using the same formulation as given by Eq.(\ref{eq:forward-1})-Eq.(\ref{eq:forward-3}). Specifically, the inference process for $l$-th layer can be formalized as:
\begin{equation}
\small
\begin{split}
    \basic{K}{\textit{test}}{l} = W_K\binput{X}{\textit{test}}{l-1},\ \basic{V}{\textit{test}}{l} = W_V\binput{X}{\textit{test}}{l-1},\ \basic{Q}{\textit{test}}{l} = W_Q\binput{X}{\textit{test}}{l-1} \\
    \binput{X}{\textit{test}}{l} = \mathscr{F}(\textbf{Attention}_l (\{ \update{K}{T}{l} \Vert \basic{K}{\textit{test}}{l} \}, \{ \update{V}{T}{l} \Vert \basic{V}{\textit{test}}{l} \}, \basic{Q}{\textit{test}}{l}))
\end{split} \numberthis
\end{equation}
In this way, we can obtain the representation $\binput{X}{\textit{test}}{L}$ produced by the final layer, which is used to make predictions.

\section{Experiments}

\newcommand{\baseline}{ICL}
\newcommand{\oursInTable}{\bv{Ours}}
\newcommand{\modelargs}[2]{\multirow{#1}{*}{\bv{#2}}}
\newcommand{\modelsizeargs}[3]{\multirow{#1}{*}{\makecell{\bv{#2}\\\texttt{#3}}}}

\newcolumntype{x}[1]{>{\centering\arraybackslash\hspace{0pt}}p{#1}}

\begin{table*}[htbp]
\centering
\small
\resizebox{1.0\linewidth}{!}{
\begin{tabular}{lx{1.2cm}x{1.0cm}x{1.0cm}x{1.0cm}x{1.3cm}c}
\toprule
\textbf{Method} &\textbf{SST2} &\textbf{SST5} &\textbf{TREC} &\textbf{MR} &\textbf{AGNews} &\textbf{Average} \\
\midrule
\multicolumn{7}{c}{\itshape In-context learning w/o dev set. $^\diamondsuit$} \\
\midrule
Random         &57.9 &27.5 &30.3 &59.5 &33.6 &41.8 \\ 
Herding~\cite{chen2010super}             &62.0 &24.8 &26.4 &54.1 &38.7 &41.2 \\
K-Center Greedy \cite{sener2018active}    &58.6 &25.1 &31.3 &59.0 &42.3 &43.3 \\
Entropy \cite{coleman2019selection}             &62.4 &25.5 &26.2 &54.1 &30.6 &39.8 \\
LeastConfidence \cite{coleman2019selection}     &58.4 &26.0 &23.5 &55.9 &31.6 &39.1 \\
Margin \cite{coleman2019selection}             &62.4 &26.1 &24.2 &54.1 &38.1 &41.0 \\
CAL  \cite{margatina-etal-2021-active}               &59.3 &25.3 &31.8 &66.2 &42.3 &45.0 \\
CRAIG  \cite{pmlr-v119-mirzasoleiman20a}             &63.4 &26.4 &32.0 &59.3 &37.4 &43.7 \\
GradMatch   \cite{killamsetty2021grad}        &57.0 &26.3 &25.8 &56.6 &32.6 &39.7 \\
FacilityLocation  \cite{iyer2013submodular}  &65.5 &23.9 &35.7 &61.7 &42.5 &45.9 \\
GraphCut  \cite{iyer2013submodular}          &65.0 &25.3 &34.7 &66.3 &41.9 &46.6 \\
\textbf{\methodname}&\textbf{85.7} &\textbf{39.2} &\textbf{54.2} &\textbf{71.6} &\textbf{72.9} &\textbf{64.7}\\
\midrule
\multicolumn{7}{c}{\itshape In-context learning w/ dev set. $^\heartsuit$} \\
\midrule
LENS \cite{li2023finding}               &86.3 &44.9 &59.0 &83.1 &77.9 &70.2 \\
Random$^\clubsuit$           &77.9 &38.0 &56.6 &81.8 &74.6 &65.8\\
\textbf{\methodname}$^\clubsuit$ &\textbf{88.1} &\textbf{45.2} &\textbf{61.6} &\textbf{84.8} &\textbf{80.3} &\textbf{72.0}\\
\bottomrule
\end{tabular}}
\caption{Experimental results across conventional ICL tasks with different ICL methods. $^\diamondsuit$ denotes that each method was assessed over ten random seeds, and the reported values are the average performance across these seeds. 
$^\heartsuit$ signifies the evaluation of multiple sets of random demonstrations on the dev set, with the best-performing set selected \cite{li2023finding}. $^\clubsuit$ indicates the methods utilized the same demonstrations, ensuring that any improvement stemmed solely from the  \methodname stage.}
\label{tab:main-result-compare}
\end{table*}

\subsection{Conventional In-context Learning Tasks}

We first evaluate the proposed \methodname against other enhanced ICL methods in a fair comparison of conventional ICL tasks. We select five popular tasks, including \textbf{SST2}~\cite{data_sst}, \textbf{SST5}~\cite{data_sst}, \textbf{MR}~\cite{data_mr}, \textbf{AGNews}~\cite{data_agnews} and \textbf{TREC}~\cite{data_trec1, data_trec2}. For a fair comparison, we choose \textbf{GPT2-L} as the base model, which is widely used by previous studies \cite{li2023finding}.

\paragraph{Compared Methods} 
We use several demonstration selection methods as baselines, which can be classified into distinct approaches. (1) Geometry-based techniques, such as Herding~\cite{chen2010super} and K-Center Greedy~\cite{sener2018active}, concentrate on spatial proximity within the feature space for constructing demonstrations. (2) Uncertainty-based methods posit that demonstrations with higher uncertainty exert more substantial influence on the model, encompassing techniques such as Entropy, Least Confidence, Margin~\cite{coleman2019selection}, and CAL~\cite{margatina-etal-2021-active}. (3) Gradient matching-based methods, such as CRAIG~\cite{pmlr-v119-mirzasoleiman20a} and GradMatch~\cite{killamsetty2021grad}, aim to replicate the gradient distribution of the full dataset with a subset. (4) Submodularity-based methods assess informativeness and diversity for selection, including such as FacilityLocation and GraphCut~\cite{iyer2013submodular}. (5) LENS~\cite{li2023finding} adopts a ``filter-then-search'' approach, utilizing the ``InfoScore'' metric to select the best demonstrations. Notably, our fairest baseline is the Random method (vanilla ICL), where we use the exact same demonstrations without any selection process. 

\paragraph{Further Comparison} 
To assess the performance of \methodname across a range of LMs across different sizes, we extend the base model of \methodname on conventional ICL tasks to include \textbf{OPT} (125M, 350M, 2.7B)~\cite{zhang2022opt}, \textbf{GPT-2} (Medium, Large, and XL)~\cite{radford2019language-gpt2}, \textbf{GPT-Neo} (2.7B)~\cite{gpt-neo} and \textbf{LLaMA2} (7B, 13B)~\cite{touvron2023llama}. This extension aims to demonstrate the effectiveness of \methodname across a spectrum of LM scales.

\subsection{Challenging Benchmarks}

In contrast to the conventional ICL benchmarks that entail a candidate pool for sample selection, real-world complex tasks frequently present scenarios where only a limited and fixed set of demonstrations is available. This particular challenge renders many existing methods of demonstration selection impractical in such scenarios.
To deal with the challenges, we choose \textbf{MMLU}~\cite{hendryckstest2021} and \textbf{BBH}~\cite{srivastava2022beyond} to extend the evaluation of \methodname to more rigorous and multifaceted scenarios. 
Concretely, MMLU encompasses a diverse set of 57 tasks, spanning elementary mathematics, US history, computer science, law, and various other domains. 
In contrast, BBH is tailored to address a suite of 23 challenging tasks within the BIG-Bench framework.
In addressing these challenging benchmarks, we employ more advanced LLMs,
including \textbf{LLaMA2}~\cite{touvron2023llama} (7B and 13B) and \textbf{Pythia}~\cite{biderman2023pythia} (70M, 410M, 1.4B, 6.9B and 12B) as base models, given their balanced ability and versatility in handling a wide range of tasks. 

\begin{table*}[htbp]
    \small
    \centering
    \resizebox{1.0\linewidth}{!}{
    \begin{tabular}{c|c|x{1.5cm}x{1.5cm}x{1.5cm}x{1.5cm}x{1.5cm}c}
        \toprule
        \bf{Model} & \multicolumn{1}{c|}{\bf{Method}} & \bf{SST2} & \bf{SST5} & \bf{TREC} & \bf{MR} & \bf{AGNews} & \bf{Average} \\
        \midrule
        \multirow{2}{*}{\makecell{\bv{OPT}-\texttt{125M}}}
        & \baseline & 55.7 & 26.7 & 25.0 & 50.4 & 41.7 & 39.9 \\
        & \oursInTable & \textbf{72.0} & \textbf{33.2} & \textbf{47.0} & \textbf{65.8} & \textbf{50.6} & \textbf{53.7}\\
        \midrule
        \multirow{2}{*}{\makecell{\bv{OPT}-\texttt{350M}}}
        & \baseline     & 54.1 & 26.6 & 37.2  & 71.2 & 42.9 & 46.4 \\
        & \oursInTable  & \textbf{79.7} & \textbf{31.8} & \textbf{45.8}  & \textbf{73.4} & \textbf{64.3} & \textbf{59.0} \\
        \midrule
        \modelsizeargs{2}{GPT2-Medium}{355M}
        & \baseline & 59.6 & 23.7 & 33.4 & 65.0 & 51.7 & 46.7 \\
        & \oursInTable & \textbf{86.9} & \textbf{38.1} & \textbf{43.6} & \textbf{80.3} & \textbf{80.0} & \textbf{65.8} \\
        \midrule
        \modelsizeargs{2}{GPT2-XL}{1.5B}
        & \baseline & 60.3 & 41.1 & 34.4 & 66.7 & 56.9 & 51.9 \\
        & \oursInTable & \textbf{89.3} & \textbf{43.8} & \textbf{60.6} & \textbf{86.1} & \textbf{82.6} &\textbf{72.5} \\
        \midrule
        \multirow{2}{*}{\makecell{\bv{OPT}-\texttt{2.7B}}}
        & \baseline    & 62.4 & 45.8 & 37.0 & 86.2 & 77.8 & 61.8 \\
        & \oursInTable & \textbf{72.4} & \textbf{47.7} & \textbf{50.0} & \textbf{89.0} & \textbf{85.8} & \textbf{69.0} \\
        \midrule
        \modelsizeargs{2}{GPT-Neo}{2.7B}
        & \baseline & 84.8 & 39.7 & 49.6 & 85.2 & 71.6 & 66.2 \\
        & \oursInTable & \textbf{88.1} & \textbf{45.5} & \textbf{59.2} & \textbf{89.1} & \textbf{83.5} & \textbf{73.1} \\
        \midrule
        \modelsizeargs{2}{LLaMA2}{7B}
        & \baseline    & 89.5 & 46.3 & 82.8 & 91.2 & 84.6 & 78.9 \\
        & \oursInTable & \textbf{90.0} & \textbf{48.1} & \textbf{84.8} & \textbf{92.2} & \textbf{88.9} & \textbf{80.8} \\
        \midrule
        \modelsizeargs{2}{LLaMA2}{13B}
        & \baseline    & 95.2 & 46.4 & 84.8 & 92.5 & 87.0 & 81.2 \\
        & \oursInTable & \textbf{96.0} & \textbf{49.9} & \textbf{86.2} & \textbf{94.7} & \textbf{88.8} & \textbf{83.1} \\
        \bottomrule
    \end{tabular}
    }
    \caption{Experimental results cross different LLMs on conventional ICL tasks. To ensure that any observed improvement stems exclusively from the \methodname stage, all variables are held constant across experiments.}
    \label{tab:result-classical}
\end{table*}

\subsection{Implementation Details and Evaluation}
All experiments are conducted on a single NVIDIA A100 GPU. 
For all baselines and \methodname, we run each method over ten random seeds and report the average performance.
For conventional ICL tasks, we follow \cite{li2023finding} that the number of demonstrations for SST2, SST5, TREC, MR, and AGNews is 8, 10, 12, 8, and 8, respectively. For MMLU and BBH, the demonstrations come from the dataset's inherent demonstrations. Specifically five demonstrations for MMLU and three demonstrations for BBH per task. 
In the in-context setting without a dev set, we fix the iteration number $T$ at $5$, with the gating parameter $\eta$ set to $0.01$. In the in-context setting with a dev set, we relax the max iteration number $T$ to $15$, using the dev set to determine the final hyper-parameters. For the dev set, we randomly select a sample size identical to the test set and keep it fixed.
For evaluation, similar to previous methods, we concatenate the test input with each candidate's answer and submit them to the LLM. The final answer is selected by summing the probabilities of the tokens belonging to the answer part and choosing the candidate answer with the highest probability.

\subsection{Main Results}

\paragraph{Results on Conventional ICL Tasks}
We first compare \methodname with previous methods on conventional ICL tasks. Table \ref{tab:main-result-compare} shows that \methodname consistently outperforms baseline methods. 
In addition, a significant improvement is observed when comparing its performance with and without the utilization of a development set.
Particularly, \methodname surpasses the Random baseline by an average margin of 6.2\% under the dev setting. This improvement is solely attributed to the iterative forward operations, providing empirical evidence of the effectiveness of the proposed method.

\paragraph{Transferablity across Different LMs}
In the aforementioned experiments, we employ the same LM (GPT-L) as the base model. To assess the transferability of \methodname across LMs of varying scales and pre-trained corpora, we expand our experimental scope to include diverse settings. Table \ref{tab:result-classical} validates the transferability and generalizability of \methodname across different base models, maintaining competitiveness even when applied to stronger models such as LLaMA2. 

\begin{table*}[htbp]
    \centering
    \resizebox{1.0\linewidth}{!}{
    \begin{tabular}{c|rr|rr|rr|rr|rr|rr|rr}
    \toprule
    \multirow{4}{*}{\makecell{\textbf{Model}\\\&\\\textbf{Method}}}
    &\multicolumn{4}{c|}{\bv{LLaMA2}} &\multicolumn{10}{c}{\bv{Pythia}} \\
    \cmidrule{2-15}
    &\multicolumn{2}{c|}{\texttt{7B}} &\multicolumn{2}{c|}{\texttt{13B}} &\multicolumn{2}{c|}{\texttt{70M}}&\multicolumn{2}{c|}{\texttt{410M}}&\multicolumn{2}{c|}{\texttt{1.4B}} &\multicolumn{2}{c|}{\texttt{6.9B}} &\multicolumn{2}{c}{\texttt{12B}}\\
    \cmidrule{2-15}
    & \bv{ICL} & \bv{Ours} & \bv{ICL} & \bv{Ours}  & \bv{ICL} & \bv{Ours}  & \bv{ICL} & \bv{Ours}  & \bv{ICL} & \bv{Ours}  & \bv{ICL} & \bv{Ours}  & \bv{ICL} & \bv{Ours}  \\
    \midrule
    \textbf{MMLU} & 41.9 & \textbf{44.6} & 45.1 & \textbf{47.6} & 24.6 & \textbf{29.6} & 27.0 & \textbf{30.8} & 30.6 & \textbf{33.6} & 33.5 & \textbf{37.1} & 36.2 & \textbf{39.5}\\
    \textbf{BBH} & 46.1 & \textbf{49.8} & 49.7 & \textbf{53.6} & 34.9 & \textbf{39.8} & 37.5 & \textbf{42.3} & 38.2 & \textbf{43.8} & 39.2 & \textbf{43.4} & 41.1 & \textbf{45.3} \\
    \bottomrule
    \end{tabular}}
    \caption{The results of vanilla ICL and \methodname on challenging benchmarks, including \textbf{MMLU} and \textbf{BBH}.}
    \label{tab:result-mmlu-bbh}
\end{table*}

\begin{figure*}[htbp]
    \centering
    \includegraphics[width=1.0\linewidth]{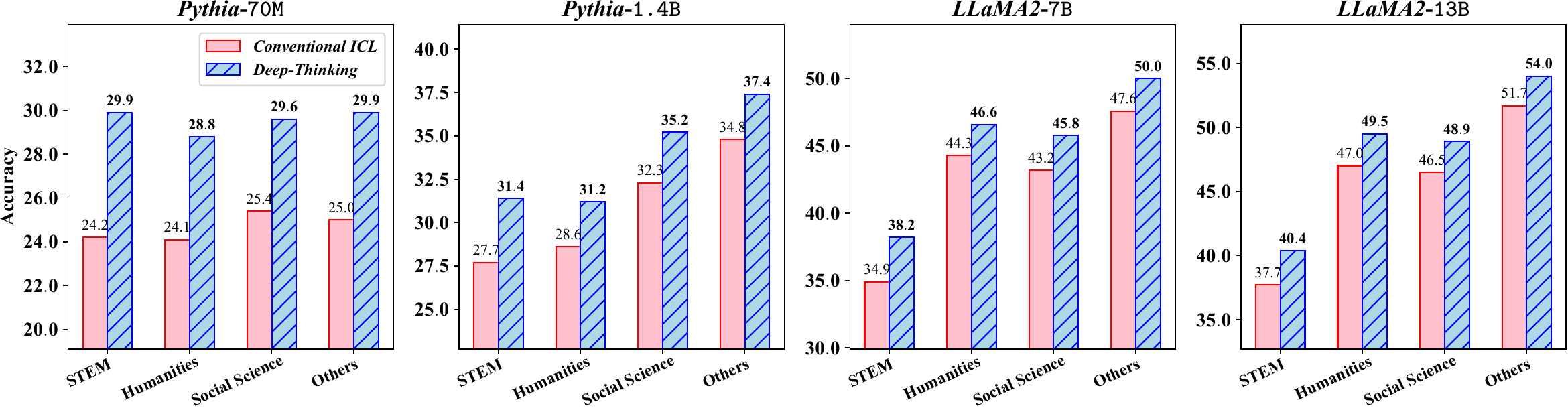}
    \caption{Comparison of model performance across four major classes of the \textbf{MMLU} benchmarks. 
    Due to space constraints and to ensure clarity in presentation, we solely report the results of four out of the seven models.}
    \label{fig:mmlu-part-improv}
\end{figure*}

\paragraph{Results on Challenging Benchmarks}
Table~\ref{tab:result-mmlu-bbh} presents the averaged performance of \methodname and vanilla ICL on MMLU and BBH benchmarks. Notably, the performance boost is consistent across all selected models, affirming that \methodname beneficially impacts a wide spectrum of frontier LMs, independent of their specific designs or training data. This effect is particularly evident in smaller-sized models such as Pythia, where the relative performance uplift is significant. This trend aligns with observations from conventional ICL tasks, further highlighting the broad applicability and effectiveness of \methodname in enhancing models' in-context learning ability.

\subsection{Fine-grained Task Analysis}
To investigate whether \methodname shows greater advantages in tasks requiring complex reasoning, we conduct a detailed analysis as shown in Figure \ref{fig:mmlu-part-improv}. The MMLU benchmark categorizes its 57 subtasks into four major classes: STEM, Humanities, Social Science, and Others. 
STEM tasks rigorously assess the model's reasoning abilities, whereas the other three categories predominantly serve as tests of knowledge retention.
STEM tasks pose greater challenges for vanilla ICL methods; however, \methodname consistently demonstrates improvements across all categories, indicating a relatively more substantial gain in STEM. For instance, LLaMA2-7B exhibits a 3.9\% increase over ICL in STEM, while registering improvements of 2.3\%, 2.6\%, and 2.5\% in Humanities, Social Science, and Others, respectively. This highlights the effectiveness of \methodname in enhancing models' capabilities to address complex reasoning and problem-solving tasks.

\subsection{Impact of Hyper-parameters}

\begin{figure}[!t]
    \centering
    \includegraphics[width=1.0\linewidth]{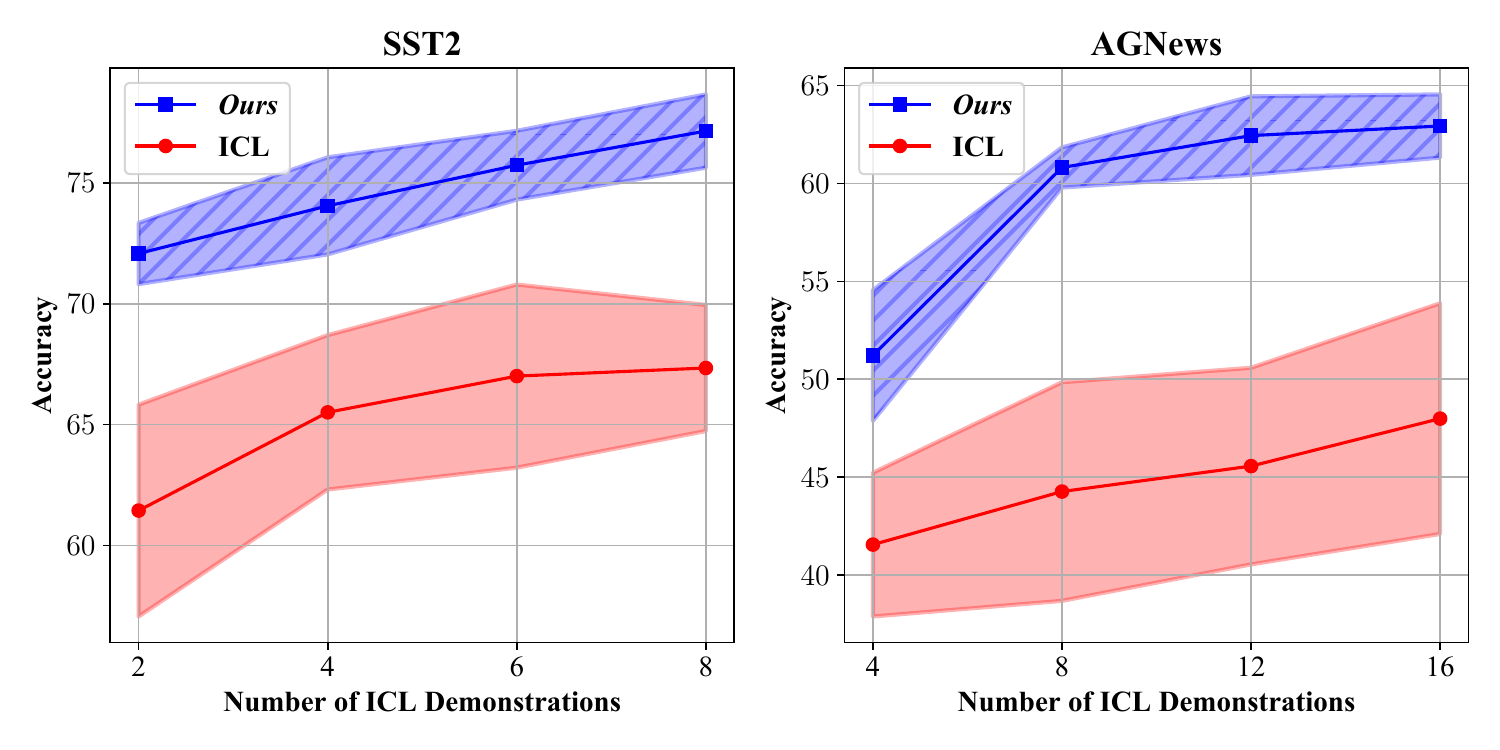}
    \caption{An illustration of the impact of increasing the number of demonstrations on the effectiveness of vanilla ICL and \methodname.}
    \label{fig:shots-randomness}
\end{figure}

\begin{figure}[htbp]
    \centering
    \includegraphics[width=1.0\linewidth]{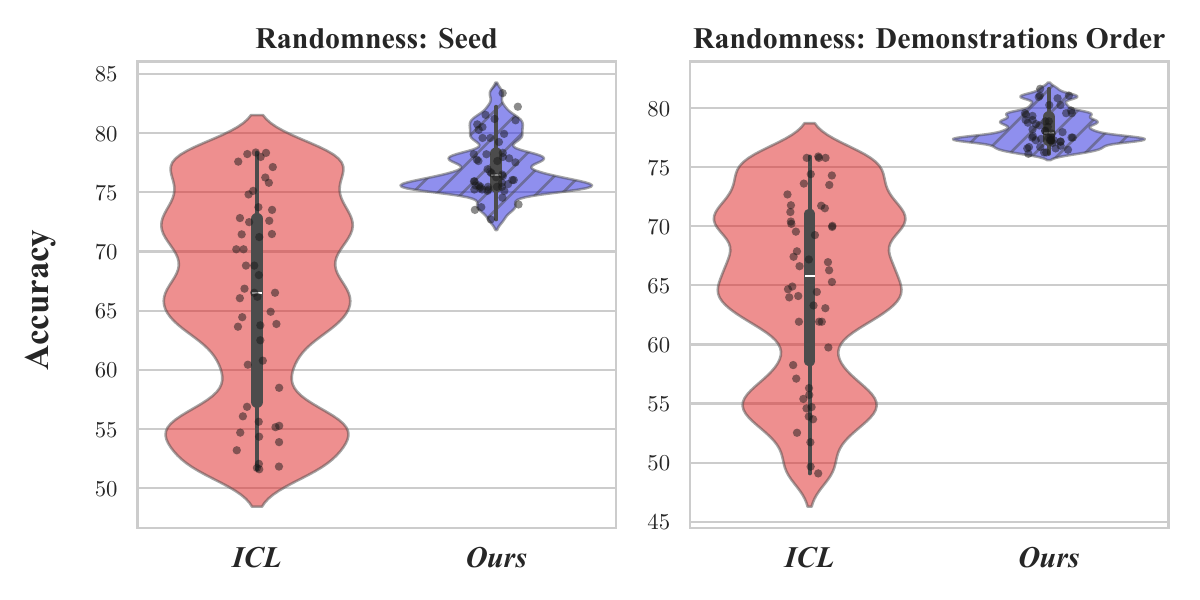}
    \caption{The performance distribution of performance for vanilla ICL and \methodname, comparing effects of random seeds (left) and random orders (right).}
    \label{fig:violin-randomness}
\end{figure}

\paragraph{Impact of Demonstration Numbers}
As depicted in Figure 4, we conducted experiments on the SST2 and AGNews datasets to explore the impact of increasing the number of demonstrations on the efficacy of ICL. 
We perform ten runs of the experiments and calculate the variance. The results indicate that both vanilla ICL and \methodname benefit from an increase in the number of demonstrations. However, \methodname consistently outperforms vanilla ICL, achieving significantly better results even with a smaller number of demonstrations. Additionally, \methodname demonstrates a smaller variance, indicating greater robustness. This suggests that it is more cost-effective to ``think'' more from existing demonstrations than merely increasing the number of demonstrations.

\paragraph{The Sensitivity of Random Seed}
The randomness in vanilla ICL and \methodname stems solely from the random selection of demonstrations.
To further investigate the robustness of the methods, we conduct multiple experiments on the SST dataset by randomly choosing eight different demonstrations, keeping other variables (except the seed) constant. As illustrated in Figure \ref{fig:violin-randomness} (left), vanilla ICL is significantly affected by randomness, whereas \methodname achieves stronger performance with less variance. 
\methodname, by iterating multiple times, bridges the gap by maximizing the utility of demonstrations.

\paragraph{The Sensitivity of Demonstrations Orders}
The order of demonstrations in ICL is crucial and can significantly impact performance~\cite{lu-etal-2022-fantastically,liu-etal-2022-makes}. In particular, different orders of demonstrations can lead to performance close to the state-of-the-art or merely random guesses. We examine the effect of demonstration order on ICL and \methodname on the SST-2 dataset. As shown in Figure~\ref{fig:violin-randomness} (right), the results show that vanilla ICL is highly sensitive to the order, with a significant variance in outcomes indicating large instability. \methodname benefits from iterative learning of demonstrations, overcoming order biases, and thus shows more consistent performance.

\begin{figure}[!t]
    \centering
    \includegraphics[width=1.0\linewidth]{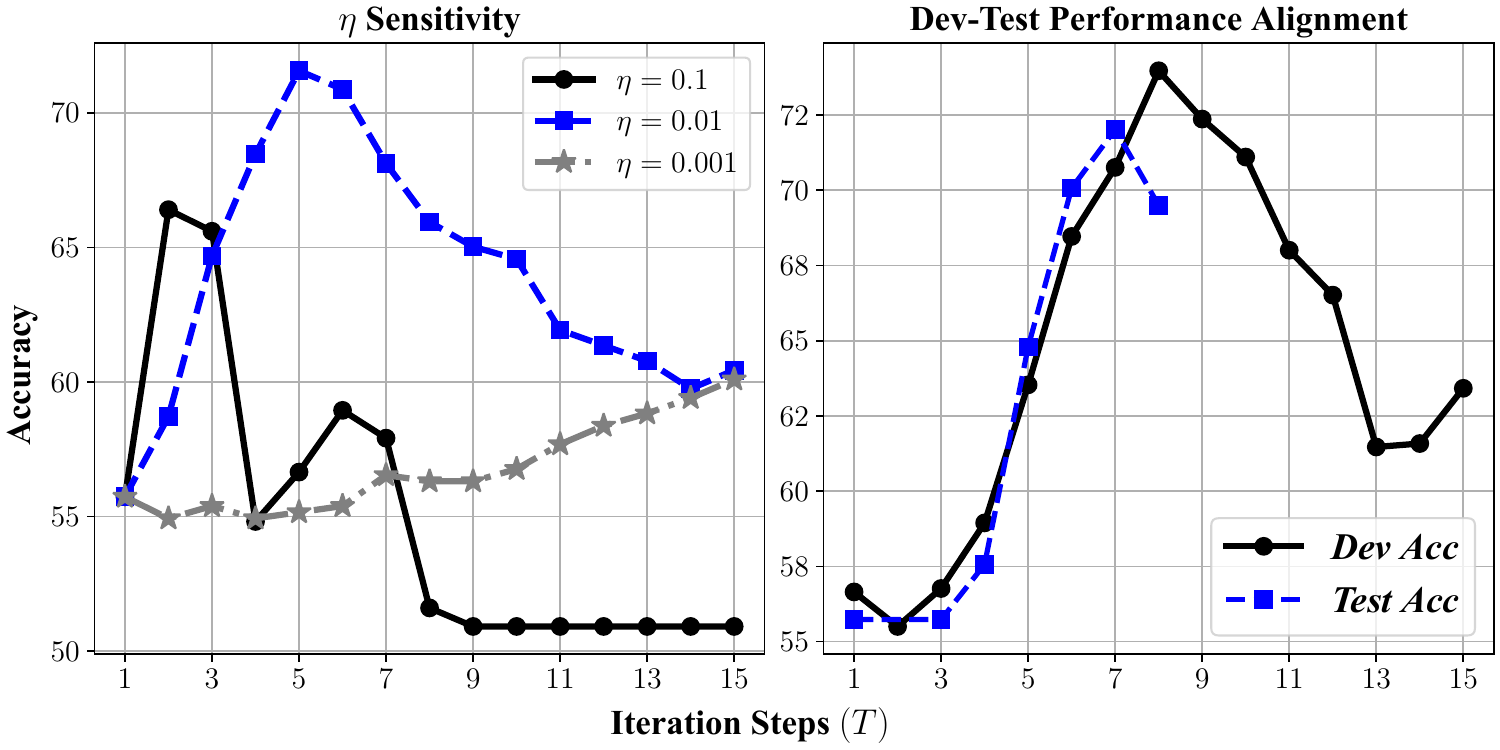}
    \caption{Sensitivity of accuracy to hyperparameter $\eta$ and alignment of development (Dev) and test set performance over iteration steps (T).}
    \label{fig:eta-impact}
\end{figure}

\paragraph{Impact of Gate $\eta$ and Iteration Steps $T$}
The gate $\eta$ is crucial in dictating how much of the memory is retained during the \methodname stage and the degree of openness to new information for the next iteration. A larger $\eta$ signifies greater changes, thus requiring fewer iterations $T$, and vice versa. To investigate the optimal $\eta$, we enumerate values in [0.001, 0.01, 0.1]. As shown in Figure \ref{fig:eta-impact} (left), setting $\eta=0.01$ achieves a balance between the number of iterations and performance. 
We can analogize $\eta$ to the learning rate and $T$ to the number of training steps. Inspired by this comparison, as described in Table~\ref{tab:main-result-compare}, we use a dev set to determine the optimal number of iterations $T$. Figure \ref{fig:eta-impact} (right) shows that there is a basic alignment between the dev and test sets.

\section{Related Work}
In-context learning (ICL) with LLMs has made a breakthrough and become mainstream in tackling various tasks~\cite{li2023can, dong2022survey, qiao2022reasoning}. Recently, great efforts have been made to improve the performance of ICL from different perspectives, such as demonstrations selection \cite{liu-etal-2022-makes,li2023finding}, prompt template design \cite{wei2021finetuned}, and intermediate chain-of-thought (CoT) reasoning \cite{wei2022chainofthought,zhang2023automatic}. 

For demonstration selection, \citet{liu-etal-2022-makes} performed demonstration selection through a $k$NN-based retriever, choosing the closest example to test input. \citet{wu2022selfadaptive} proposed self-adaptive ICL with a general select-and-rank framework for demonstration selection. In addition to example selection, \citet{yao2022fantastically} investigated the sensitivity of ICL to the permutation of demonstrations and proposed entropy metrics to determine their order. The above ICL methods are usually restricted by the number of demonstrations. To mitigate such a challenge, \citet{hao2022structured} attempted to scale ICL by grouping demonstrations, which could increase the number of demonstrations to 1,000. 

The formatting function also plays a crucial role in ICL, especially for tasks requiring complex reasoning steps, such as commonsense reasoning. \citet{wei2022chainofthought} introduced chain-of-thoughts (CoT) prompting to provide guidance. \citet{zhang2023automatic} stimulated the model's ability for gradual reasoning by adding the ``\textit{Let's think step-by-step}'' prefix. Instead of generating reasoning steps,  \citet{press2022measuring} investigated the compositional reasoning abilities by allowing LLMs to generate follow-up questions. Subsequently, \citet{madaan2023selfrefine} introduced a new framework to enhance the initial outputs generated by LLMs via iterative feedback and refinement. 
Meanwhile, some studies \cite{xie2021explanation,dai2022gpt,wang2023large} attempt to uncover the underlying working mechanism of ICL. In particular, \citet{xie2021explanation} showed that ICL happened via Bayesian inference, where certain concepts were implicitly predicted before the final prediction. Subsequently, \citet{dai2022gpt} revealed that there are connections between ICL and explicit fine-tuning and explained LLMs as meta-optimizers \cite{irie2022dual}. 

Unlike existing methods, to the best of our knowledge, we are the first to decouple ICL into two stages and focus on how to deeply learn from fixed demonstrations rather than on demonstration selection or prompt engineering. This is advantageous for the world situation where provided samples are scarce, i.e., there is no large candidate set of demonstrations.

\section{Conclusion} 

In this paper, we introduce a novel two-stage framework aimed at enhancing the ICL capabilities of LLMs by leveraging iterative forward inferences to learn from demonstrations. By decoupling the ICL process into a dedicated \methodname stage for demonstration training and a subsequent test stage, we effectively mimic the decision-making process of humans by learning from analogy. This approach aligns with how humans engage in repeated logical thinking. 
The empirical evaluations across conventional ICL benchmarks and more challenging datasets demonstrate that our \methodname strategy significantly outperforms previous ICL approaches, particularly in scenarios where demonstration selection is impractical.

\section*{Limitations} 
While our method has demonstrated promising results and significant advancements across various aspects, it is imperative to conduct a thorough analysis of its limitations. In this section, we explore the potential constraints of our method. Firstly, owing to limited computational resources and time constraints, we were unable to evaluate our method on larger language models, such as LLaMA2-70B. This limitation may impact the generalizability of our findings to larger-scale language models. Secondly, our evaluation primarily focused on conventional ICL tasks and challenging benchmarks. To enhance the comprehensiveness of our findings, we intend to broaden the scope of evaluation to encompass a diverse range of dataset types, including math reasoning, code generation, and open-ended text generation. This extension aims to provide further validation of our method's generalizability.

\section*{Acknowledgments}
Min Yang was supported by National Key Research and Development Program of China (2022YFF0902100), National Natural Science Foundation of China (62376262), the Natural Science Foundation of Guangdong Province of China (2024A1515030166), Shenzhen Science and Technology Innovation Program (KQTD20190929172835662), Shenzhen Basic Research Foundation (JCYJ20210324115614039).
This work was supported by Alibaba Group through Alibaba Research Intern Program.

\bibliography{custom}

\clearpage

\appendix

\onecolumn

\section{Data Statistics and Templates For In-context Learning Tasks}

We choose five datasets for evaluating in-context learning methods following \cite{min-etal-2022-noisy, li2023finding}. We show the prompting templates and dataset statistics in Table~\ref{tab:task_template_cls}.

\newcommand{\simpleCat}[2]{\texttt{#1: \{query\}}\newline{}\texttt{#2: \{label\}}}

\makeatletter
\newcommand{\joinList}[1]{\texttt{#1}\checknextarg}
\newcommand{\checknextarg}{\@ifnextchar\bgroup{\gobblenextarg}{}}
\newcommand{\gobblenextarg}[1]{\texttt{ / }\texttt{#1}\@ifnextchar\bgroup{\gobblenextarg}{}}

\begin{table}[htbp]
    \centering
    \resizebox{1.0\linewidth}{!}{
    \begin{tabular}{ccc|p{0.25\linewidth}p{0.5\linewidth}}
        \toprule
        \bv{Task}   & \bv{Original Dev Size} & \bv{Test Size} & \bv{Template}                 & \bv{Labels} \\
        \midrule
        \bv{SST2}   & 67349 & 873 & \simpleCat{Review}{Sentiment} & \joinList{negative}{positive}     \\
        \midrule
        \bv{SST5}   & 8544 & 2210 & \simpleCat{Review}{Sentiment} & \joinList{terrible}{negative}{neutral}{positive}{great}                \\
        \midrule
        \bv{TREC}   & 5452 & 500 & \simpleCat{Question}{Type}    & \joinList{Abbreviation}{Entity}{Description}{Person}{Location}{Number} \\
        \midrule
        \bv{MR}     & 8530 & 1066 & \simpleCat{Review}{Sentiment} & \joinList{negative}{positive}                                          \\
        \midrule
        \bv{AGNews} & 120000 & 7601 & \simpleCat{Article}{Category} & \joinList{World}{Sports}{Business}{Technology}                         \\
        \bottomrule
    \end{tabular}}
    \caption{The statistics of standard in-context learning tasks, including detailed task sizes, prompting templates, and labels.}
    \label{tab:task_template_cls}
\end{table}

\section{Data Statistics For MMLU}

We obtained the MMLU dataset from the Hugging Face Hub, specifically from the repository \texttt{cais/mmlu}\footnote{\url{https://huggingface.co/datasets/cais/mmlu}}. According to the dataset's card, MMLU encompasses 57 tasks spanning diverse knowledge domains. Each task includes a minimum of 100 test examples. For in-context demonstrations, five examples per task provided by original dataset are used. We present detailed statistics for each sub-task, including the classification scheme, in Table~\ref{tab:task_mmlu_stat}.

\begin{table}[htbp]
    \centering
    \begin{tabular}{p{1.0\linewidth}}
        \toprule
        \textbf{STEM} \\
        \textbf{astronomy}: 152, \textbf{college\_physics}: 102, \textbf{conceptual\_physics}: 235, \textbf{high\_school\_physics}: 151,\\ \textbf{college\_chemistry}: 100, \textbf{high\_school\_chemistry}: 203, \textbf{college\_biology}: 144,\\
        \textbf{high\_school\_biology}: 310, \textbf{college\_computer\_science}: 100, \textbf{computer\_security}: 100,\\
        \textbf{high\_school\_computer\_science}: 100, \textbf{machine\_learning}: 112, \textbf{abstract\_algebra}: 100,\\
        \textbf{college\_mathematics}: 100, \textbf{elementary\_mathematics}: 378, \textbf{high\_school\_mathematics}: 270,\\
        \textbf{high\_school\_statistics}: 216, \textbf{electrical\_engineering}: 145 \\
        \midrule
        \textbf{Humanities} \\
        \textbf{high\_school\_european\_history}: 165, \textbf{high\_school\_us\_history}: 204, \textbf{high\_school\_world\_history}: 237,\\
        \textbf{prehistory}: 324, \textbf{formal\_logic}: 126, \textbf{logical\_fallacies}: 163, \textbf{moral\_disputes}: 346,\\
        \textbf{moral\_scenarios}: 895, \textbf{philosophy}: 311, \textbf{world\_religions}: 171, \textbf{international\_law}: 121,\\
        \textbf{jurisprudence}: 108, \textbf{professional\_law}: 1534 \\
        \midrule
        \textbf{Social Sciences} \\
        \textbf{high\_school\_government\_and\_politics}: 193, \textbf{public\_relations}: 110, \textbf{security\_studies}: 245,\\
        \textbf{us\_foreign\_policy}: 100, \textbf{human\_sexuality}: 131, \textbf{sociology}: 201, \textbf{econometrics}: 114,\\
        \textbf{high\_school\_macroeconomics}: 390, \textbf{high\_school\_microeconomics}: 238, \\
        \textbf{high\_school\_geography}: 198, \textbf{high\_school\_psychology}: 545, \textbf{professional\_psychology}: 612 \\
        \midrule
        \textbf{Others (Business, Health, misc.)} \\
        \textbf{global\_facts}: 100, \textbf{miscellaneous}: 783, \textbf{professional\_accounting}: 282, \textbf{business\_ethics}: 100,\\
        \textbf{management}: 103, \textbf{marketing}: 234, \textbf{anatomy}: 135, \textbf{clinical\_knowledge}: 265,\\
        \textbf{college\_medicine}: 173, \textbf{human\_aging}: 223, \textbf{medical\_genetics}: 100, \textbf{nutrition}: 306,\\
        \textbf{professional\_medicine}: 272, \textbf{virology}: 166 \\
        \bottomrule
    \end{tabular}
    \caption{The number of samples for each subtask in the \textbf{MMLU} benchmark.}
    \label{tab:task_mmlu_stat}
\end{table}

\section{Data Statistics For BBH}

For the BigBench-Hard (BBH) dataset, we sourced the data from the \texttt{maveriq/bigbenchhard}\footnote{\url{https://huggingface.co/datasets/maveriq/bigbenchhard}}. We have excluded certain tasks due to their incompatibility with multiple-choice or classification formats. Specifically, the tasks omitted include: Dyck Languages, Multistep Arithmetic, Object Counting, Word Sorting, and Reasoning about Colored Objects. Table~\ref{tab:task_bbh_stat} provides a detailed statistics for each selected task.

\begin{table}[htbp]
    \centering
    \begin{tabular}{p{1.0\linewidth}}
        \toprule
        \textbf{BBH} \\
        \textbf{boolean\_expressions}: 250, \textbf{causal\_judgement}: 187, \textbf{date\_understanding}: 212,\\
        \textbf{disambiguation\_qa}: 247, \textbf{formal\_fallacies}: 250, \textbf{geometric\_shapes}: 200,\\
        \textbf{hyperbaton}: 250, \textbf{logical\_deduction\_three\_objects}: 250, \textbf{logical\_deduction\_five\_objects}: 250,\\
        \textbf{logical\_deduction\_seven\_objects}: 250, \textbf{movie\_recommendation}: 231, \textbf{navigate}: 250,\\
        \textbf{penguins\_in\_a\_table}: 145, \textbf{ruin\_names}: 247, \textbf{salient\_translation\_error\_detection}: 250,\\
        \textbf{snarks}: 177, \textbf{sports\_understanding}: 250, \textbf{temporal\_sequences}: 250, \textbf{web\_of\_lies}: 250 \\
        \textbf{tracking\_shuffled\_objects\_three\_objects}: 250, \textbf{tracking\_shuffled\_objects\_five\_objects}: 250,\\
        \textbf{tracking\_shuffled\_objects\_seven\_objects}: 250\\
        \bottomrule
    \end{tabular}
    \caption{The number of samples for each selected subtask in the \textbf{BBH} benchmark.}
    \label{tab:task_bbh_stat}
\end{table}

\clearpage

\section{Minimal Implementation}

The provided minimal implementation code showcases our proposed method, primarily integrating with the Hugging Face Transformers library.

\begin{minipage}[h]{\linewidth}
\begin{lstlisting}[language=PythonPlus]
class GatedOptim:
    def __init__(self, step_size=0.01):
        self.step_size = step_size

    def upd_x(self, old_x, g):  return old_x + self.step_size * g
    def __call__(self, old_xs, new_xs):
        pesudo_gs = [new_x - old_x for old_x, new_x in zip(old_xs, new_xs)]
        updated_kv = [self.upd_x(old_x, g) for old_x, g in zip(old_xs, pesudo_gs)]
        return updated_kv

class AttnOptimWrapper:
    def __init__(self, llm, **optimizer_args):
        self.model = llm
        self.kv = None
        self.update_k = GatedOptim(**optimizer_args)
        self.update_v = GatedOptim(**optimizer_args)

    def step(self, ctx_ids):
        L = len(ctx_ids)
        ctx_ids = ctx_ids.unsqueeze(0)  # [1, L]
        mask = torch.ones_like(ctx_ids).repeat(1, 2 if self.kv else 1)

        out = self.model(ctx_ids, mask, past_key_values=self.kv, use_cache=True)
        out_kv = out.past_key_values  # kv @ (old_ctx + new_ctx)
        cur_kv = [[k[:,:,-L:,:], v[:,:,-L:,:]] for k, v in out_kv]  # kv @ (new_ctx)

        if not self.kv:
            self.kv = cur_kv
        else:
            (old_ks, old_vs), (cur_ks, cur_vs) = zip(*self.kv), zip(*cur_kv)
            upd_ks = self.update_k(old_ks, cur_ks)
            upd_vs = self.update_v(old_vs, cur_vs)
            self.kv = list(zip(upd_ks, upd_vs))  # kv @ (merged_ctx)
        return self.kv

def main():
    # ... initialization (dataset, model, tokenizer, logger, etc)
    ex_str = load_exemplar()
    ex_ids, ex_mask = tokenize(ex_str)
    meta_optim = AttnOptimWrapper(model)
    for idx in range(args.meta_steps):
        ex_kv = meta_optim.step(ex_ids)
        for B_query_ids, B_query_mask in inference_loader:
            bs = len(B_query_ids)
            # [B(expanded), L+L'(concat)]
            B_merged_mask = torch.cat((ex_mask.expand(bs, -1), B_query_mask), dim=1)
            B_kv_shape = (bs, -1, -1, -1)
            B_kv = [[k.expand(B_kv_shape), v.expand(B_kv_shape)] for k, v in ex_kv]

            B_out = model(B_query_ids, B_merged_mask, past_key_values=B_kv).logits
            B_out = F.log_softmax(B_out, dim=-1)
            # ...
\end{lstlisting}
\end{minipage}

\end{document}